\documentclass{article}
\usepackage{ijcai20}

\usepackage{times}
\usepackage{soul}
\usepackage{url}
\usepackage[hidelinks]{hyperref}
\usepackage[utf8]{inputenc}
\usepackage[small]{caption}
\usepackage{graphicx}
\usepackage{amsmath}
\usepackage{amsthm}
\usepackage{booktabs}
\usepackage{algorithm}
\usepackage{multirow}
\usepackage{amsfonts}
\usepackage{algpseudocode}
\usepackage{color}

\newcommand{\nop}[1]{}
\newcommand{\model}{\textbf{StrGNN}~}
\newcommand{\modele}{\textbf{StrGNN}}

\title{Structural Temporal Graph Neural Networks\\ for Anomaly Detection in Dynamic Graphs}

\author{
Lei Cai$^{1,2,}$\thanks{Work done during an internship at NEC Labs America.}\and
Zhengzhang Chen$^{2,}$\thanks{Corresponding author: \{zchen\}@nec-labs.com.}\and
Chen Luo$^3$\and
Jiaping Gui$^2$\and\\
Jingchao Ni $^2$\and
Ding Li$^2$\and
Haifeng Chen$^2$ 
\affiliations
$^1$Washington State University, USA\\
$^2$NEC Laboratories America, USA\\
$^3$Rice University, USA\\
}

\begin{document}

\maketitle

\begin{abstract}
Detecting anomalies in dynamic graphs is a vital task, with numerous practical applications in areas such as security, finance, and social media. Previous network embedding based methods have been mostly focusing on learning good node representations, 
whereas largely ignoring the subgraph structural changes related to the target nodes in dynamic graphs. 
In this paper, we propose \modele, an end-to-end structural temporal Graph Neural Network model for detecting anomalous edges in dynamic graphs. In particular, we first extract the $h$-hop enclosing subgraph centered on the target edge and propose the node labeling function to identify the role of each node in the subgraph. Then, we leverage graph convolution operation and Sortpooling layer to extract the fixed-size feature from each snapshot/timestamp. Based on the extracted features, we utilize Gated recurrent units (GRUs) to capture the temporal information for anomaly detection. Extensive experiments on six benchmark datasets and a real enterprise security system demonstrate the effectiveness of \modele. 
\end{abstract}

\section{Introduction}


Recent studies of dynamic graphs/networks have witnessed a growing interest. Such dynamic graphs model a variety of systems including societies, ecosystems, the Internet, and others. For example, 
in enterprise dynamic network~\cite{luo2018tinet}, the node represents a system entity (such as process, file, and Internet sockets) and an edge indicates the corresponding interaction between two system entities. \nop{Other examples include world wide web connecting websites across the world, and social networks that connect users, businesses, or customers using relationships such as hyperlinks, friendship, or collaborations.} These dynamic networks, unlike static networks, are constantly changing. Possible changes include graph structure change or modification of node attributes.

A fundamental task on dynamic graph analysis is anomaly detection---identifying objects, relationships, or subgraphs, whose ``behaviors'' significantly deviate from underlying majority of the network~\cite{ICDE.2011.5767885,Ranshous:2015}. In this work, we focus on the anomalous edge detection in dynamic graphs. 
Detecting anomalous edges can help understand the system status and diagnose system fault \cite{Ranshous:2015,akoglu2015graph}. For example, in an enterprise dynamic network, some system entity pairs, such as a user software and system-specific internet socket ports (\textit{e.g.}, port number $\leq 1024$), never form an edge (interaction/connection) in-between in normal system environments. Once occurring, these suspicious interactions/activities may indicate some serious cyber-attack happened and could significantly damage the enterprise system~\cite{cheng2016ranking}. 

\begin{figure}[t]
    \includegraphics[width=\columnwidth]{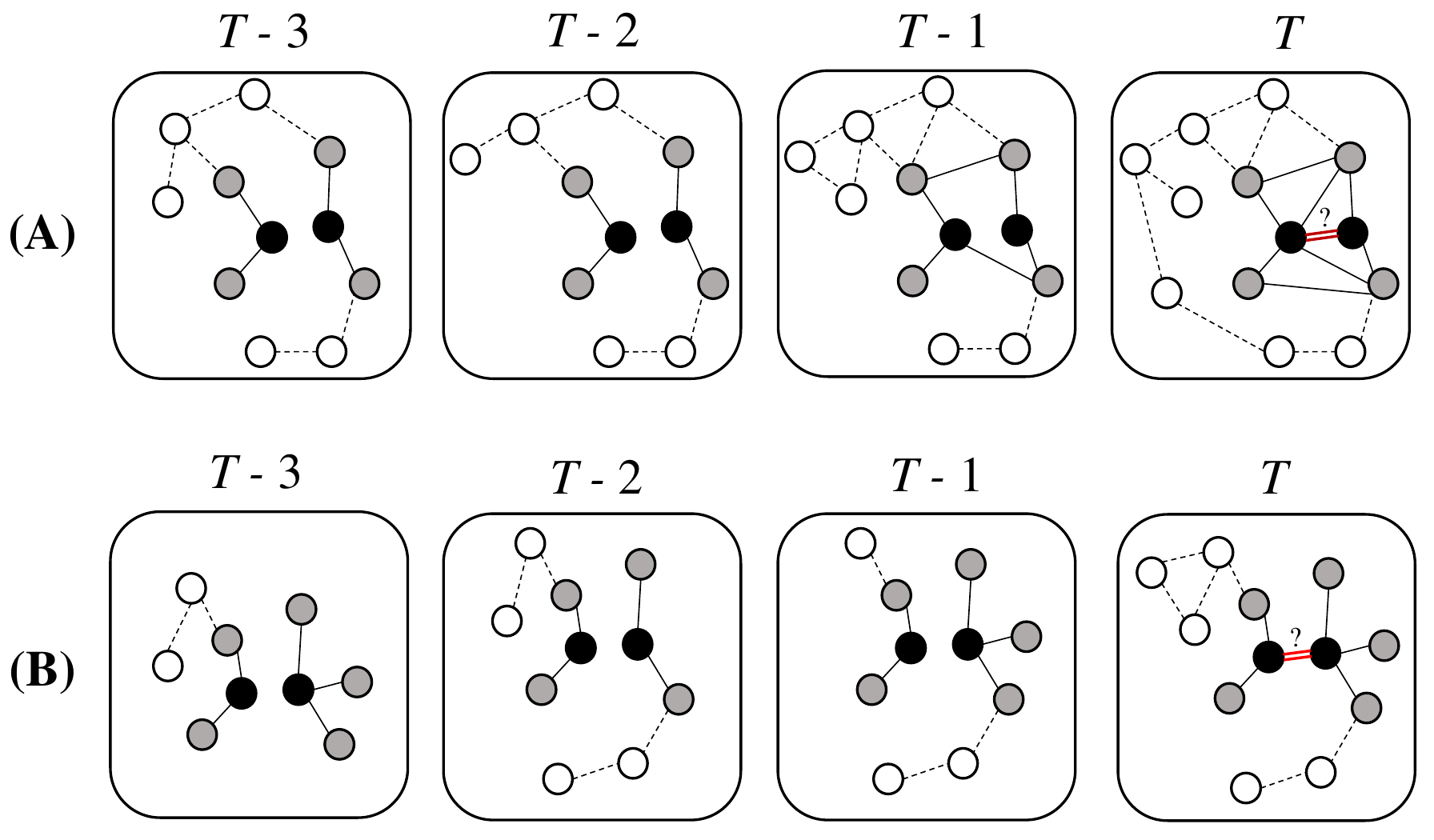}
	\caption{An example of structural changes in dynamic graphs.} \label{fig:intro}
\end{figure}

Recently, graph embedding has shown to be a powerful tool in learning the low-dimensional representations in networks that can capture and preserve the graph structure. However, most existing graph embedding approaches are designed for static graphs, and thus may not be suitable for a dynamic environment in which the network representation has to be constantly updated. Only a few advanced embedding-based methods (such as NetWalk \cite{yu2018netwalk}) are suitable for updating the representation dynamically as the network evolves. However, these methods require the knowledge of the nodes over the whole time span and thus can hardly promise the performance on new nodes in the future. More importantly, these methods neglect a notable characteristic of the dynamic networks---the subgraph structural changes related to the target nodes.
These structural temporal dynamics are key to understanding system behavior. For example, in Figure ~\ref{fig:intro}, the target edge at timestamp $t$ is marked as a double red line, and the $1$-hop subgraph centered on the target edge is marked with gray. It can be seen from Figure \ref{fig:intro} (A) that the interactions between nodes of the subgraph (\textit{i.e.}, gray nodes) become more frequent. Therefore, the target edge in Figure \ref{fig:intro} (A) is reasonable to be a normal edge. In contrast, in Figure \ref{fig:intro} (B), there is no interactions between the neighbors of the subgraph from timestamp $t-3$ to $t-1$. Therefore, the target edge at timestamp $t$ is more likely to be an anomalous edge. Thus, it is critical to model and detect the structural changes over time for the anomaly detection task.

To address the aforementioned issues, we propose \modele, a \underline{str}uctural \underline{g}raph \underline{n}eural \underline{n}etwork to identify anomalous edges in dynamic graphs. 
\model is designed to detect unusual subgraph structures centered on the target edge in a given time window while considering the temporal dependency. 
\model consists of three sub-models: \textbf{ESG} (Enclosing Subgraph Generation), \textbf{GSFE} (Graph Structural Feature Extraction), and \textbf{TDN} (Temporal Detection Network).
First, \textbf{ESG} extracts a $h$-hop enclosing subgraph centered on the target edge from each graph snapshot. Subgraphs extracted based on different edges can result in the same topology structure. Thus, a node labeling function is proposed to indicate the role of each node in the subgraph. 
Then, \textbf{GSFE} module leverages Graph Convolution Neural Network and pooling technologies to extract fixed-size feature from each subgraph. Based on the extracted features, 
\textbf{TDN} employs the Gated recurrent units (GRUs) to capture the temporal dependency for anomaly detection. Different from the previous embedding based methods, the whole process of \model can be trained end-to-end, \textit{i.e.}, \model takes the test edges along with the original dynamic graphs as input and directly outputs the category (\textit{i.e.}, anomaly or normal) for each test edge. Moreover, our proposed \model framework focuses on mining the structural temporal patterns in a given time window. Therefore, node embedding is not required to learn and \model is not sensitive to the edge and vertex changes (such as new nodes) in the dynamic graphs. 
We conduct extensive experiments on six benchmark datasets to evaluate the performance of \modele. The results demonstrate the effectiveness of our proposed algorithm. We also apply \model to a real enterprise security system for intrusion detection. By using \modele, we can reduce false positives of the state-of-the-art methods by at least 50\%, while keeping zero false negatives.

\section{Related Work}
In this section, we briefly introduce previous work on embedding based anomaly detection in graphs.  

\nop{\subsection{Anomaly Detection}
	Many traditional machine learning methods have been proposed to tackle anomaly detection tasks. One-class SVM \cite{scholkopf2000support} was proposed to learn the boundary of normal samples, and detect the anomaly based on the learned boundary. Due to the computation cost, it takes more time on the large scale dataset. Robust covariance \cite{pena2001multivariate} fits the data with a pre-defined distribution, and detect the anomaly by computing the distance between the sample and estimated distribution. Isolation forest \cite{liu2008isolation}  was proposed to isolate and detect the anomaly. Local outliers detection \cite{breunig2000lof} estimates the density of a given dataset and detects the anomaly. When the feature space of a given dataset can represent the relationship between samples, the density-based method is efficient to detect the anomaly. Anomaly detection is more challenging in graph setting due to the complexity of the data. Lots of work were proposed to learn the network embedding of the graph, which can further be employed to detect anomaly combing with the tradition anomaly detection method. }

\subsection{Anomaly Detection on Static Graphs}
Inspired by word embedding methods~\cite{mikolov2013distributed} 
in natural language processing tasks, recent advances such as DeepWalk~\cite{perozzi2014deepwalk}, LINE~\cite{tang2015line}, and Node2Vec~\cite{grover2016node2vec} have been proposed to learn node embedding via the \textbf{skip-gram} technology. The DeepWalk generates random walks for each vertex with a given length and picks the next step uniformly from the neighbors. 
Different from DeepWalk, the LINE~\cite{tang2015line} preserves not only the first-order (observed tie strength) relations but also the second-order proximities (shared neighborhood structures of the vertices). Node2Vec~\cite{grover2016node2vec} uses two different sampling strategies (breadth-first sampling and depth-first sampling) for vertices that result in different feature representations. Through the network embedding technology, both anomalous node and edge detection tasks can be performed with traditional anomaly detection methods.

\subsection{Anomaly Detection on Dynamic Graphs}
Dynamic graphs are more complex due to the variation of the graph structure. That is, the vertices and edges are changing along the time dimension. 
To capture the dependency between different graphs along the time dimension, recently few network embedding based methods have been proposed~\cite{zhou2018dynamic}. Dyngem \cite{goyal2018dyngem} employs the auto-encoder method to learn the embedding for each graph, and a constraint loss function is employed to minimize the difference between all graphs. Dyngraph2vec~\cite{goyal2019dyngraph2vec} uses the Recurrent Neural Network to capture the temporal information and learn the embedding using auto-encoder technology. Recently, NetWalk~\cite{yu2018netwalk}, one of the state-of-the-art methods for anomaly detection in dynamic networks, is proposed to learn the embedding while considering the temporal dependency and detect the anomaly using the density-based method. The NetWalk generates several random walks for each vertex and learns a unified embedding for each node using auto-encoder technology. The embedding representation is updated along the time dimension. 

\section{Method}
In this section, we introduce our method in detail. We start with the overall framework of our proposed Structural Temporal Graph Neural Networks for anomaly detection in dynamic graphs. The details of each component in our proposed method are introduced afterwards.

\subsection{Overall Framework}
\begin{figure*}[t]
	\centering
	\includegraphics[width=0.9\textwidth]{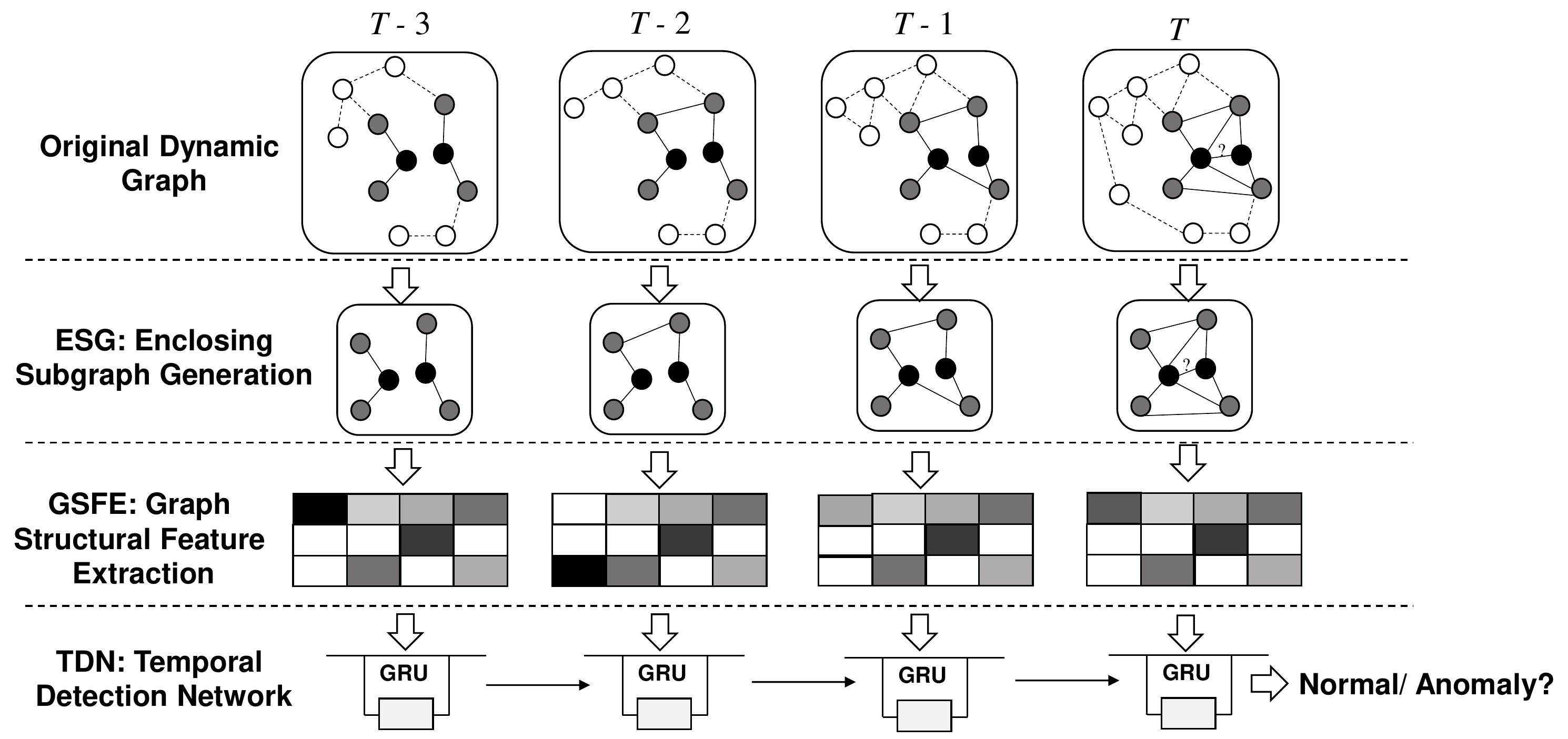}
	\caption{Illustration of our proposed \model framework.} \label{fig:frame}
\end{figure*}

Compared with the anomaly detection in a static graph, dynamic graphs are more complex and challenging in two perspectives: (1) The anomalous edges cannot be determined by the graph from a single timestamp. The detection procedure must take the previous graphs into consideration; (2) Both the vertex and edge sets are changing over time. To tackle these challenges, we propose \modele, a structural temporal Graph Neural Network framework. The key idea of our proposed method is to capture structural changes centered on the target edge in a given time window and determine the category (\textit{i.e.}, anomaly or normal) of the target edge based on the structural changes. Our proposed \model framework consists of three key components: \textbf{ESG} (Enclosing Subgraph Generation), \textbf{GSFE} (Graph Structural Feature Extraction), and \textbf{TDN} (Temporal Detection Network), as illustrated in Figure \ref{fig:frame}.

\nop{As the first step of our proposed \model algorithm, \textbf{ESG} module generates the structural patterns related to the target edge in a time window with length of $w$. Since the detection is most relevant to the local subgraph, we employ the $h$-hop enclosing subgraph centered on each target edge to preserve graph structure and enable a more efficient computation. In addition, the $h$-hop enclosing subgraph can help alleviate the impacts from noisy data. The \textbf{ESG} module generates $h$-hop enclosing subgraph for each timestamp in the detection window. Then, the \textbf{GSFE} module takes these subgraphs as the input and leverages the graph pooling technology to generate fixed-size feature for each subgraph. Since the anomalous edges must be determined by considering the historical graphs, as the last step, \textbf{TDN} module employs Gated recurrent units (GRUs) to capture the temporal information. GRUs take the graph structural feature from the current timestamp and temporal feature from the previous timestamp as the input and generate the temporal feature. The temporal feature from the last timestamp is employed to predict anomalous edges using a neural network. We introduce the details of each component in the following sections.}

\subsection{\textbf{ESG}: Enclosing Subgraph Generation}
For the first module, Enclosing Subgraph Generation, our goal is to generate enclosing subgraph structure related to the target edge so as to detect the anomalies more efficiently. Directly employing the whole graph for analysis can be highly computational expensive, especially considering the real-world networks with thousands or even millions of nodes and edges. Recent work~\cite{xu2018representation} also proved that in Graph Neural Networks, each node is most influenced by its neighbors.   

Definition 1. (Enclosing subgraph in static graphs) For a static network $G=(V, E)$, given a target edge $e$ with source node $x$ and destination node $y$, the $h-$hop enclosing subgraph $G_{x,y}^h$ centered on edge $e$ can be obtained by $\{i|d(i,x) \leq h \lor d(i,y) \leq h\}$, where $d(i,x)$ is the shortest path distance between node $i$ and node $x$. 

Definition 2. (Enclosing subgraph in dynamic graphs) For a temporal network $\{G(i)=\{V(i), E(i)\}\}_{i=t-w}^t$ with window size $w$, given a target edge $e^t$ with source node $x^t$ and destination node $y^t$, the $h-$hop enclosing subgraph $G_{x_t,y_t}^h$ centered on edge $e^t$ is a collection of all subgraph centered on $e^t$ in the temporal network $\{G(i)_{x_t, y_t}^h | t-w \leq i \leq t \}$.


For a target edge $e^t$, we extract the enclosing subgraph in dynamic graphs based on Definition 2. 
However, the extracted subgraph only contains topological information. Subgraphs extracted based on different edges can result in the same topological structure. 
To distinguish the role of each node in the subgraph, in this work, we propose to annotate the nodes in the subgraph with different labels. 
A good node labeling function should convey the following information: 1) which edge is the target edge in the current subgraph, and 2) the contribution of each node in identifying the category of each edge. More specifically, given the edge $e^t$ and the corresponding source and destination node $x^t$ and $y^t$, our node labeling function for the enclosing subgraph $G(i)_{x_t, y_t}^h$ is defined as follows:
\begin{eqnarray} \label{eq: label}
f(i, x_t, y_t) = 1+ \min(d(i,x^t),d(i,y^t)) \\ \nonumber
+ (d_{sum} / 2)[(d_{sum}/2) + (d_{sum} \% 2) -1] ,
\end{eqnarray}
where $d(i,x^t)$ is the shortest path distance between node $i$ and node $x^t$, and $d_{sum}=d(i,x^t)+d(i,y^t)$. In addition, the two center nodes are labeled with 1. If a node $i$ satisfies $d(i,x^t)=\infty$ or $d(i,y^t)=\infty$, it will be labeled as 0. The label will be converted into a one-hot vector as the attribute $X$ for each node. By employing the node labeling function, we can generate the label for each node, which can represent structure information for the given subgraph. The category of the target edge $e^t$ at timestamp $t$ can be predicted by analyzing the labeled subgraph in the given time window.

\subsection{\textbf{GSFE}: Graph Structural Feature Extraction}

To analyze the structure of each enclosing subgraph from the given time period, the Graph Convolution Neural Network (GCN)~\cite{kipf2016semi} can be employed to project the subgraph into an embedding space. In GCN, the graph convolution layer was proposed to learn the embedding of each node in the graph and aggregate the embedding from its neighbors. The layer-wise forward operation of graph convolution layer can be described as follows:
\begin{equation}
G(X,A) = \sigma(\hat{D}^{-1/2}\hat{A}\hat{D}^{-1/2}XW),
\end{equation}
where $\hat{A} = A + I $ is the summation of the adjacency matrix and identity matrix, $\sigma (\cdot)$ denotes an activation function, such as the $ReLU(\cdot) = max(0, \cdot)$, and $W$ is the trainable weight matrix. By employing the graph convolution layer, each node can aggregate the embedding from its neighbors. By stacking the graph convolution layer in the neural network, each node can obtain more information from other nodes. For example, each node can obtain information from its $2$-hop neighbors by stacking two graph convolution layers. 

GCN can generate node embedding for detecting anomalous edges in a single graph. However, in our dynamic graph setting, the anomalies should be determined in the context of $\{G(i)_{x_t, y_t}^h | t-w \leq i \leq t \}$. The number of nodes in different enclosing subgraphs is commonly different, thus results in different sizes of the feature vector in different subgraphs. Therefore, it is challenging to analyze the dynamic graphs using Graph Neural Networks due to the various sizes of the input. 

To tackle this problem, we leverage the graph pooling technology to extract the fixed-size feature for each enclosing subgraph. Any graph pooling method can be employed in our proposed \model framework to extract the fixed-size feature for further analysis. In this work, we employ the Sortpooling layer proposed by \cite{zhang2018end}, which can sort the nodes in the enclosing subgraph based on their importance and select the feature from the top $K$ nodes.

Given the node embedding $H_i$ corresponding to graph $G(i)_{x_t, y_t}^h$, the importance score for each node in the Sortpooling layer is defined as follows:
\begin{equation}\label{score}
S(H_i, A) = \sigma(\hat{D}^{-1/2}\hat{A}\hat{D}^{-1/2}H_i W^1), 
\end{equation}
where $A$ is the adjacency matrix of graph $G(i)_{x_t, y_t}^h$, and $W^1$ is the projection matrix with output channel 1. Each node can obtain the importance score by using Equation~\ref{score}. All nodes in the enclosing subgraph will be sorted in order of the importance score. And only the top $K$ nodes will be selected for further analysis. For the subgraphs that contain less than $K$ nodes, the zero-padding will be employed to guarantee that each subgraph contains the same fixed-size feature.

\subsection{\textbf{TDN}: Temporal Detection Network}
\label{subsec:tdn}
The Graph Structural Feature Extraction module can generate low-dimensional features for anomaly detection. However, it does not consider the temporal information, which is of great importance for determining the category (\textit{i.e.}, anomaly or normal) of an edge in the dynamic setting. 


Given the extracted structural feature $\{\hat{H_i}\}_{i=t-w}^t$,  $H_i \in R^{K \times d}$, where $K$ is the number of selected nodes in each graph, and $d$ is the dimension of feature for each node, in this work, we employ the Gated recurrent units (GRUs) \cite{chung2014empirical}, which can alleviate the vanishing and exploding gradient problems \cite{goodfellow2016deep}, to capture the temporal information as:
\begin{eqnarray}
z_t &=& \sigma(W_z\hat{H}_t+U_zh_{t-1}+b_z)\\
r_t &=& \sigma(W_r\hat{H}_t+U_rh_{t-1}+b_r)\\
h_t^{'} &=& \tanh(W_h\hat{H}_t+U_h(r_t\circ h_{t-1})+b_h)\\
h_t &=& z_t\circ h_{t-1} + (1-z_t)\circ h_t^{'},
\end{eqnarray}
where $\circ$ represents the element-wise product operation, $W$, $U$, and $b$ are parameters. The GRU network takes the feature at each timestamp as input, and feeds the output of current timestamp into the next timestamp. Therefore, the temporal information can be modeled by the GRU network. The output of last timestamp $h_t$ is employed to analyze the category of the target edge $e^t$. The anomalous edge detection problem can be formulated as follows:
\begin{equation} \label{eq:loss}
L = -(y^t\log(g(h_t)) + (1-y^t)\log(1-g(h_t))),
\end{equation}
where $g(\cdot)$ is a fully connected network, and $y^t$ is the category of edge $e^t$.

For the anomaly detection task, in many real-world cases, the dataset does not contain any anomalous samples or only contain a small number of anomalous samples. 

One straightforward way of generating negative samples is to draw samples from a ``context-independent'' noise distribution (such as Random sampling or injected sampling \cite{akoglu2015graph}), where a negative sample is independently and does not depend on the observed samples. 
However, due to the large anomalous edge space, this noise distribution would be very different from the data distribution, which would lead to poor model learning. Thus, in this work, we propose ``context-dependent'' negative sampling strategy. 

The intuition behind our strategy is to generate negative samples from ``context-dependent'' noise distribution. Here, the ``context-dependent'' noise distribution for the sampled data $E'$ is defined as: $P_{E'} \sim P(E)*(\frac{1}{N*|E|})$, where $P(E)$ denotes the observed data distribution, $|E|$ is the number of edges in the graph, and $N$ is the number of nodes in the graph. Specifically, we first randomly sample a normal vertex pair $ e= \{x_a,x_b\}$ in the graph. Then, we replace one of the nodes, say $x_a$ with a randomly sampled node $x'$ in the graph and form a new negative sample $e' = \{x',x_b\}$. If $e'$ is not belongs to the normal graph, we retain the sample, otherwise, we delete it. 

The proposed \model framework is quite flexible and easy to be customized. Any network that can capture the temporal information can be used in our proposed framework, such as Convolution Neural Network (CNN) and Vanilla Recurrent Neural Network (RNN).


\nop{
\begin{algorithm}[t]
	\caption{\modele: Structural Graph Neural Networks for Anomaly Detection in Dynamic Graphs}\label{algo:train}
	\begin{algorithmic}[1]
		\floatname{algorithm}{Algorithm}
		\renewcommand{\algorithmicrequire}{\textbf{Input:}}
		\renewcommand{\algorithmicensure}{\textbf{Output:}}
		\algnewcommand\AND{\textbf{and }}
		
		\Require $\{G(t)=\{V(t), E(t)\}\}_{t=1}^n$, the number of hops $h$, and window size $w$. 
		
		\State Inject anomalous edges if required;
		\State Extract $h$-hop enclosing subgraphs based on Definition 2;
		\State Generate the label for each node in the subgraph using Equation \ref{eq: label};
		\State Extract the graph structure features;
		\State Model the temporal information using GRU;
		\State Train the classifier using Equation \ref{eq:loss}.
		
	\end{algorithmic}
\end{algorithm}
}

\section{Experiments}
In this section, we evaluate \model on six benchmark datasets and a real enterprise network.

\subsection{Datasets}
We conduct experiments on six public datasets from different domains. 
The UCI Messages dataset \cite{opsahl2009clustering} is collected from an online community platform of students at the University of California, Irvine. Each node in the constructed graph represents a user in the platform. And the edge indicates that there is a message interaction between two users. The Digg dataset \cite{de2009social} is collected from a news website digg.com. Each node represents a user of the website, and each edge represents a reply between two users. The Email dataset is a dump of emails of Democratic National Committee. Each node corresponds to a person. And the edge indicates an email communication between two persons. The Topology \cite{zhang2005collecting} dataset is the network connections between autonomous systems of the Internet. Nodes are autonomous systems, and edges are connections between autonomous systems. The Bitcoin-alpha and Bitcoin-otc \cite{kumar2016edge,kumar2018rev2} datasets are collected from two Bitcoin platform named Alpha and OTC, respectively. Nodes represent users from the platform. If one user rates another user on the platform, there is an edge between them.

\nop{
\begin{table}[t]
	\caption{Statistics of the datasets in accumulated graph setting.}
	\label{tbl:his}
	\centering
	\resizebox{0.45\textwidth}{!}{
		\begin{tabular}{lccccc}
			\hline
			Dataset & \#Vertex & \#Edge  & \#Timestamp \\
			\hline
			\hline
			UCI Messages & 1,899 & 13,838  & 190 \\
			Digg & 30,360 & 85,155 & 16  \\
			Email & 2,029 & 3,724 & 20 \\
			Topology & 34,761 & 107,661 & 21 \\
			Bitcoin-alpha & 3,783 & 14,124 & 63 \\
			Bitcoin-otc & 5,881 & 21,492 & 63 \\ 
			\hline
		\end{tabular}
	}
\end{table}

\begin{table}[t]
	\caption{Statistics of the datasets in time-evolving graph setting.}
	\label{tbl:envolving}
	\centering
	\resizebox{0.45\textwidth}{!}{
		\begin{tabular}{lccccc}
			\hline
			Dataset & \#Vertex & \#Edge  & \#Timestamp \\
			\hline
			\hline
			UCI Messages & 1,899 & 59,835  & 190 \\
			Digg & 30,360 & 87,627 & 16  \\
			Email & 2,029 & 39,264 & 20 \\
			Topology & 34,761  & 171,403 & 21 \\
			Bitcoin-alpha & 3,783 & 24,186 & 63 \\
			Bitcoin-otc & 5,881 & 35,592 & 63 \\ 
			\hline
		\end{tabular}
	}
\end{table}
}
\begin{table}[t]
	\caption{AUC results with different hops of enclosing subgraph on UCI Messages.}
	\label{tbl:uci}
	\centering
	\begin{tabular}{lccc}
		\hline
		& 1\% & 5\% & 10\%\\
		\hline
		\hline
		$1$-hop enclosing subgraph & 0.8179 & 0.8252 & 0.7959\\
		$2$-hop enclosing subgraph & 0.8216 & 0.8274 & 0.7987\\
		$3$-hop enclosing subgraph & 0.8227 & 0.8294 & 0.8005\\
		\hline
	\end{tabular}
\end{table}

\begin{table*}[ht]
	\caption{AUC comparison on benchmark datasets.}
	\label{tbl:rh}
	\centering
	\resizebox{\textwidth}{!}{
		\begin{tabular}{l|ccc|ccc|ccc}
			\hline
			\multirow{2}{*}{Methods} & \multicolumn{3}{c|}{UCI} & \multicolumn{3}{c|}{Digg}& \multicolumn{3}{c}{Email}  \\
			
			&    1\%   &   5\%    &    10\% &    1\%   &   5\%    &    10\%&    1\%   &   5\%    &    10\% \\ \hline
			Node2Vec &  0.7371     &  0.7433     &  0.6960   &   0.7364    &   0.7081    & 0.6508 &    0.7391   &  0.7284  & 0.7103   \\
			Spectral Clustering &  0.6324     &  0.6104     &  0.5794   &   0.5949    &   0.5823    & 0.5591 &    0.8096  &  0.7857     & 0.7759 \\
			DeepWalk &  0.7514     &  0.7391     &  0.6979   &   0.7080    &   0.6881    & 0.6396 &    0.7481   &  0.7303  &  0.7197  \\
			NetWalk &  0.7758     &  0.7647     &  0.7226   &   0.7563    &   0.7176    & 0.6837 &    0.8105   &  0.8371  & 0.8305 \\
			\modele &  \textbf{0.8179}     &  \textbf{0.8252}     &  \textbf{0.7959}   &   0.8162    &   0.8254    & 0.8272 &   \textbf{ 0.8775}   &  \textbf{0.9103}     & \textbf{0.9080}   \\
			\hline 
			\multirow{2}{*}{Methods} & \multicolumn{3}{c|}{Bitcoin-Alpha}& \multicolumn{3}{c|}{Bitcoin-otc}& \multicolumn{3}{c}{Topology}  \\
			
			&    1\%   &   5\%    &    10\% &    1\%   &   5\%    &    10\%&    1\%   &   5\%    &    10\% \\ \hline
			
			Node2Vec &  0.6910     &    0.6802   & 0.6785 &  0.6951     &   0.6883    & 0.6745 &  0.6821  &  0.6752  &  0.6668  \\
			Spectral Clustering &  0.7401     &    0.7275   & 0.7167 &  0.7624     &   0.7376    & 0.7047 &   0.6685    &  0.6563     &  0.6498  \\
			DeepWalk  &  0.6985     &    0.6874   & 0.6793 &  0.7423     &   0.7356    & 0.7287 &  0.6844  &  0.6793  &  0.6682  \\
			NetWalk & 0.8385     &    0.8357   & 0.8350 &  0.7785     &  0.7694  & 0.7534 &   0.8018    &  0.8066  &  0.8058  \\
			\modele &  \textbf{0.8574}     &    \textbf{0.8667}   & \textbf{0.8627} &  \textbf{0.9012}  & \textbf{0.8775}  & \textbf{0.8836} & \textbf{0.8553}    &  \textbf{0.8352}  &  \textbf{0.8271}  \\
			\hline
		\end{tabular}
	}
\end{table*}

\nop{
\begin{table*}[ht]
	\caption{AUC comparison on time-evolving graph setting.}
	\label{tbl:te}
	\centering
	\resizebox{\textwidth}{!}{
		\begin{tabular}{l|ccc|ccc|ccc}
			\hline
			\multirow{2}{*}{Methods} & \multicolumn{3}{c|}{UCI} & \multicolumn{3}{c|}{Digg}& \multicolumn{3}{c}{Email}  \\
			
			&    1\%   &   5\%    &    10\% &    1\%   &   5\%    &    10\%&    1\%   &   5\%    &    10\% \\ \hline
			Structrual GNN + CNN &  \textbf{0.8179} & 0.8128 & 0.8109 & 0.7387 & 0.7564 & 0.7613 & 0.949 & 0.9577 & 0.9558   \\
			Structrual GNN + RNN &  0.8158 & \textbf{0.8252} & \textbf{0.8243} & 0.7561 & \textbf{0.7583} & 0.7562 & 0.9466 & 0.9574 & \textbf{0.9563} \\
			Structrual GNN + GRU &  0.8119 & 0.8225 & 0.8224 & \textbf{0.7564} & 0.7582	& \textbf{0.7590} & \textbf{0.9481} & \textbf{0.9581} & 0.9552 \\
			
			\hline 
			\multirow{2}{*}{Methods} & \multicolumn{3}{c|}{Bitcoin-Alpha}& \multicolumn{3}{c|}{Bitcoin-otc}& \multicolumn{3}{c}{Topology}  \\
			
			&    1\%   &   5\%    &    10\% &    1\%   &   5\%    &    10\%&    1\%   &   5\%    &    10\% \\ \hline
			Structrual GNN + CNN & 0.8537 & 0.8318 &   0.8485 & 0.8745 & 0.8603 & 0.8698 & 0.7716 & 0.7695 & 0.7900 \\
			Structrual GNN + RNN & 0.8352 &	0.8265 & 0.8412 & 0.8727 & 0.8586 & 0.8689 & 0.7736 & 0.772 & 0.7922  \\
			Structrual GNN + GRU & \textbf{0.8541} & \textbf{0.8290} & \textbf{0.8537} & \textbf{0.8765} & \textbf{0.8611} & \textbf{0.8700} & \textbf{0.7877} & \textbf{0.7826} & \textbf{0.8028} \\
			\hline
		\end{tabular}
	} 
\end{table*}
}

\begin{figure*}[t]
	\centering
	\vspace{-4pt}
	\includegraphics[width=\textwidth]{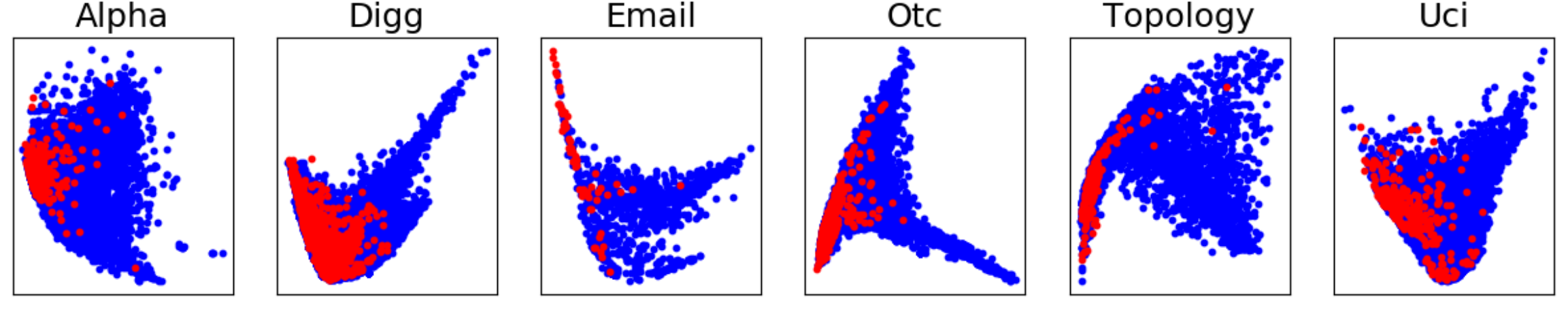}
	
	\caption{The embeddings generated by \modele. Red dots are anomalies and blue ones are normal samples.} \label{fig:emb}
	\vspace{-5pt}
\end{figure*}

\subsection{Baselines}
We compare \model with four network embedding based baselines. 

\textbf{DeepWalk}~\cite{perozzi2014deepwalk}: DeepWalk generates the random walks with given length starting from a node and learns the embedding using Skip-gram. 

\textbf{Node2Vec}~\cite{grover2016node2vec}: Node2Vec combines breadth-first traversal and depth-first traversal in the random walks generation procedure. The embedding is learned using Skip-gram technology. 

\textbf{Spectral Clustering}~\cite{spectralclustering2007}: To preserve the local connection relationship, the spectral embedding generates the node embedding by maximizing the similarity between nodes in the neighborhood.

\textbf{NetWalk}~\cite{yu2018netwalk}: NetWalk generates several random walks for each vertex and learns a unified embedding for each node using auto-encoder technology. The embedding representation will be updated along the time dimension.

framework, using an extended temporal GCN with an attention-based GRU.


For the first three baselines, after representation learning, the same K-means clustering based method~\cite{yu2018netwalk} (as in NetWalk) is used for anomaly detection. 

\subsection{Experiment Setup}

The parameters of \model can be tuned by 5-fold cross-validation on a rolling basis. Here, by default, we set the window size $w$ to $5$ and the number of hops $h$ in enclosing subgraph to $1$. We evaluate the influence of each parameter. The AUC results of \model with different $h$ on UCI Messages are shown in Table \ref{tbl:uci}. \model with $2$-hop or $3$-hop subgraph achieves similar performance as $1$-hop but requiring way more computational cost. The parameter $w$ (with $w \geq 5$) shares similar influence as $h$ on the performance of \modele. 
We employ a Graph Neural Network with three graph convolution layers to extract graph features. The size of the output feature map is set to $32$ for all three layers. The outputs of all three layers are concatenated as the embedding feature. The selected rate in the Sortpooling layer is set to $0.6$. In terms of the temporal neural network, the hidden size of GRU is set to $256$. We employ Adam method \cite{kingma2014adam} to train the network. The learning rate of Adam is set to $1e-4$. We employ batch training in the experiments and the batch size is set to $32$ for our proposed \model method. \model is end-to-end trained for $50$ epochs. We use the first $50\%$ edges as the training dataset, and the rest as the test dataset. Since the anomalous edges do not exist in the six benchmark datasets, we follow the approach used in~\cite{yu2018netwalk} to inject $1\%$, $5\%$, $10\%$ anomalous edges in the test dataset to evaluate the performance of each model. The metric used to compare the performance of different methods is AUC (the area under the ROC curve). 
The higher AUC value indicates the high quality of the method.

\subsection{Results on Benchmark Datasets}
\label{sec: syn}

\begin{figure}[t]
	\centering
	\includegraphics[width=0.4\textwidth]{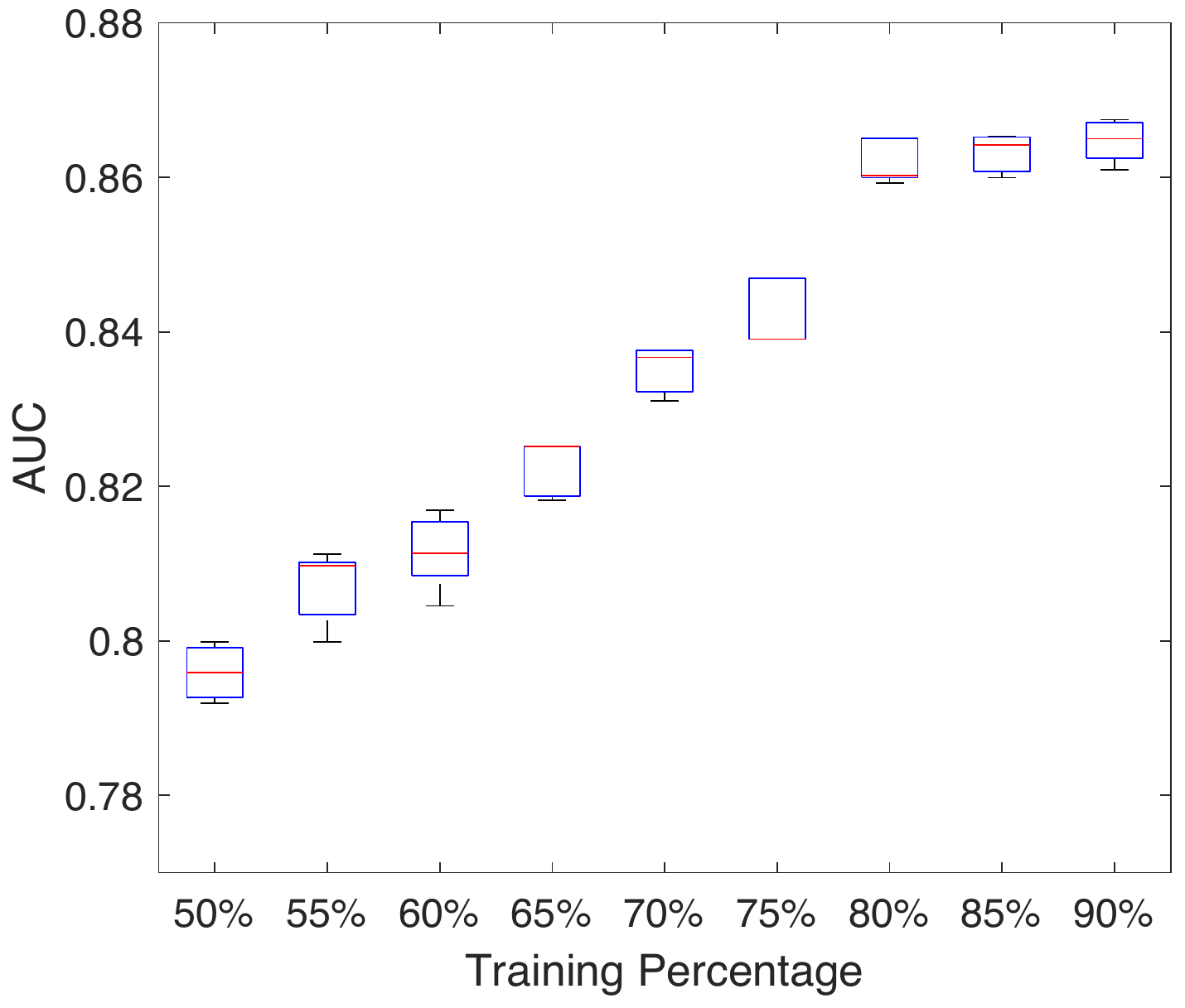}
	\caption{Stability over the training percentage of \model on UCI Messages with $10\%$ anomalies.}
	\label{fig:auc}
\end{figure}

We first compare \model with the baseline methods on six benchmark datasets. 
The experimental results in Table \ref{tbl:rh} show that \model outperforms all four baseline methods on all the benchmark datasets. And even if 10\% anomalies are injected, the performance of \model is still acceptable. 
This outstanding effect proves that \model can exploit the structural and temporal features effectively and the learned representation of the dynamic graph structure is well suited for the anomaly detection task. 

To further demonstrate the effectiveness of our \model method, we visualize the output embeddings from the GRU network of \modele. The embeddings are projected into two-dimensional space using the PCA method. The visualization results in Figure \ref{fig:emb} show that the anomalies can be easily detected using the embeddings generated by our proposed method.

In the experiments, we also evaluate our proposed model using training data with different ratios. The AUC results on UCI Messages are shown in Figure \ref{fig:auc}. It can be seen from the results that the AUC increases with the percentage of training data ranging from $50\%$ to $75\%$, and then the performance stays relatively stable.

\subsection{Intrusion Detection Application} 

To evaluate the effectiveness of \model on practical applications with real anomalies, we apply it to detect malware attacks in the enterprise environment.
We collect a $4$-week period of data from a real enterprise network composed of $109$ hosts ($87$ Windows hosts and $22$ Linux hosts). In total, there are about ten thousand normal network event records and $82$ attack records by executing $9$ different types of attacks including ATP attacks, Trojan attacks, and Puishing Email attacks at different periods. Based on the network event data, we construct an accumulated graph per day with nodes representing hosts and edges representing the network connection relationships. Based on the constructed graphs, we apply \model and the baseline methods to detect the attacks.

The AUC results are shown in Table \ref{tbl:asi}. We can see that \model achieves an increase of $9\%-28\%$ in AUC over the four baseline methods. Based on the experimental results, we also find that with the optimal hyperparameter setting, \model can capture all $82$ true alerts, while the baseline methods can only capture $72$ true alerts at most. Meanwhile, \model only generates $164$ false positives while the baseline methods generate at least $335$ false positives. The results demonstrate the effectiveness of \model in solving real-world anomaly detection tasks. 


\begin{table}[t]
	\caption{Results on intrusion detection.}
	\label{tbl:asi}
	\centering
	\begin{tabular}{lc}
		\hline
		Method & AUC \\
		\hline
		\hline
		Node2Vec & 0.71  \\
		DeepWalk & 0.76   \\
		Spectral Clustering & 0.75  \\
		Netwalk & 0.90   \\
		\modele & \textbf{0.99}  \\
		\hline
	\end{tabular}
\end{table}

\section{Conclusion}
In this paper, we investigated an important and challenging problem of anomaly detection in dynamic graphs. Different from network embedding based methods that focus on learning good node representations, we proposed \modele, a structural temporal Graph Neural Network to detect anomalous edges by mining the unusual temporal subgraph structures. \model can be trained end-to-end and it is not sensitive to the percentage of anomalies. 
We evaluated the proposed framework using extensive experiments on six benchmark datasets. The experimental results convince us of the effectiveness of our approach. We also applied \model to a real enterprise security system for intrusion detection. Our method achieved superior detection performance with zero false negatives.

\bibliographystyle{named}
\bibliography{refshort}

\end{document}